\documentclass{article}
\usepackage{ijcai22}

\usepackage{times}

\usepackage{soul}
\usepackage{url}
\usepackage[hidelinks]{hyperref}
\usepackage[utf8]{inputenc}
\usepackage[small]{caption}
\usepackage{graphicx}
\usepackage{amsmath}
\usepackage{booktabs}
\usepackage{multirow}

\newcommand*{\img}[1]{%
    \raisebox{-.02\baselineskip}{%
        \includegraphics[
        height=\baselineskip,
        width=\baselineskip,
        keepaspectratio,
        ]{#1}%
    }%
}

\newcommand*{\imgg}[1]{%
    \raisebox{-.02\baselineskip}{%
        \includegraphics[
        height=0.85\baselineskip,
        width=\baselineskip,
        keepaspectratio,
        ]{#1}%
    }%
}

\title{A Prompting-based Approach for Adversarial Example Generation and Robustness Enhancement}

\author{
    Yuting Yang\textsuperscript{\rm 1,2},
    Pei Huang\textsuperscript{\rm 2,3},
    Juan Cao\textsuperscript{\rm 1,2},
    Jintao Li\textsuperscript{\rm 1},
    Yun Lin\textsuperscript{\rm 4},
    Jin Song Dong\textsuperscript{\rm 4},
    Feifei Ma\textsuperscript{\rm 2,3,5},\\
    Jian Zhang\textsuperscript{\rm 2,3}
\affiliations
    $^1$ Key Lab of Intelligent Information Processing,\\
    Institute of Computing Technology, Chinese Academy of Sciences, Beijing, China\\
    $^2$ University of Chinese Academy of Sciences, Beijing, China\\
    $^3$ State Key Laboratory of Computer Science,\\
     Institute of Software, Chinese Academy of Sciences (ISCAS), Beijing, China\\
    $^4$ National University of Singapore, Singapore \\
    $^5$ Laboratory of Parallel Software and Computational Science, ISCAS, Beijing, China\\
\emails
\{yangyuting, caojuan\}@ict.ac.cn, \{huangpei, maff, zj\}@ios.ac.cn, \{dcsliny, dcsdjs\}nus.edu.sg
}

\begin{document}

\maketitle

\begin{abstract}
Recent years have seen the wide application of NLP models in crucial areas such as finance, medical treatment, and news media, raising concerns of the model robustness and vulnerabilities. In this paper, we propose a novel prompt-based adversarial attack to compromise NLP models and robustness enhancement technique. We first construct malicious prompts for each instance and generate adversarial examples via mask-and-filling under the effect of a malicious purpose. Our attack technique targets the inherent vulnerabilities of NLP models, allowing us to generate samples even without interacting with the victim NLP model, as long as it is based on pre-trained language models (PLMs). Furthermore, we design a prompt-based adversarial training method to improve the robustness of PLMs. As our training method does not actually generate adversarial samples, it can be applied to large-scale training sets efficiently. The experimental results show that our attack method can achieve a high attack success rate with more diverse, fluent and natural adversarial examples. In addition, our robustness enhancement method can significantly improve the robustness of models to resist adversarial attacks. Our work indicates that prompting paradigm has great potential in probing some fundamental flaws of PLMs and fine-tuning them for downstream tasks.

\end{abstract}

\section{Introduction}

Deep NLP models have shown the vulnerability to adversarial examples which are constructed by adding some imperceptible perturbations to the original input. These perturbations usually have no substantial effect on semantics but can subvert models' correct predictions. Such vulnerability has been exposed in many NLP tasks including text classification \cite{textfooler}, machine translation \cite{ZhangZC020}, dependency parsing \cite{ZhengZZHCH20}, reading comprehension \cite{LinZD20} and dialogue systems \cite{ChengWH19}. As deep learning-based models are increasingly used in safety-critical applications, the robustness issues have drawn a lot of attention in academia and industry. 

Generating and analyzing adversarial examples can help researchers better understand the robustness issues of deep NLP models and construct trustworth NLP systems. In recent years, a series of adversarial attack methods have been proposed, ranging from character-level word misspelling \cite{EbrahimiLD18}, word-level substitution \cite{seme} to sentence-level paraphrasing \cite{WangPPCWL20}. Different from that in the image field, attacks that result in illegal text sentences can be easily detected by spelling and grammar error correction \cite{PruthiDL19}. However, the attack methods based on search and rules lack naturalness and diversity in adversarial example generation which only explores a limited subset of the whole adversarial sample space.
\begin{figure*}
		\centering
		\includegraphics[scale=0.49]{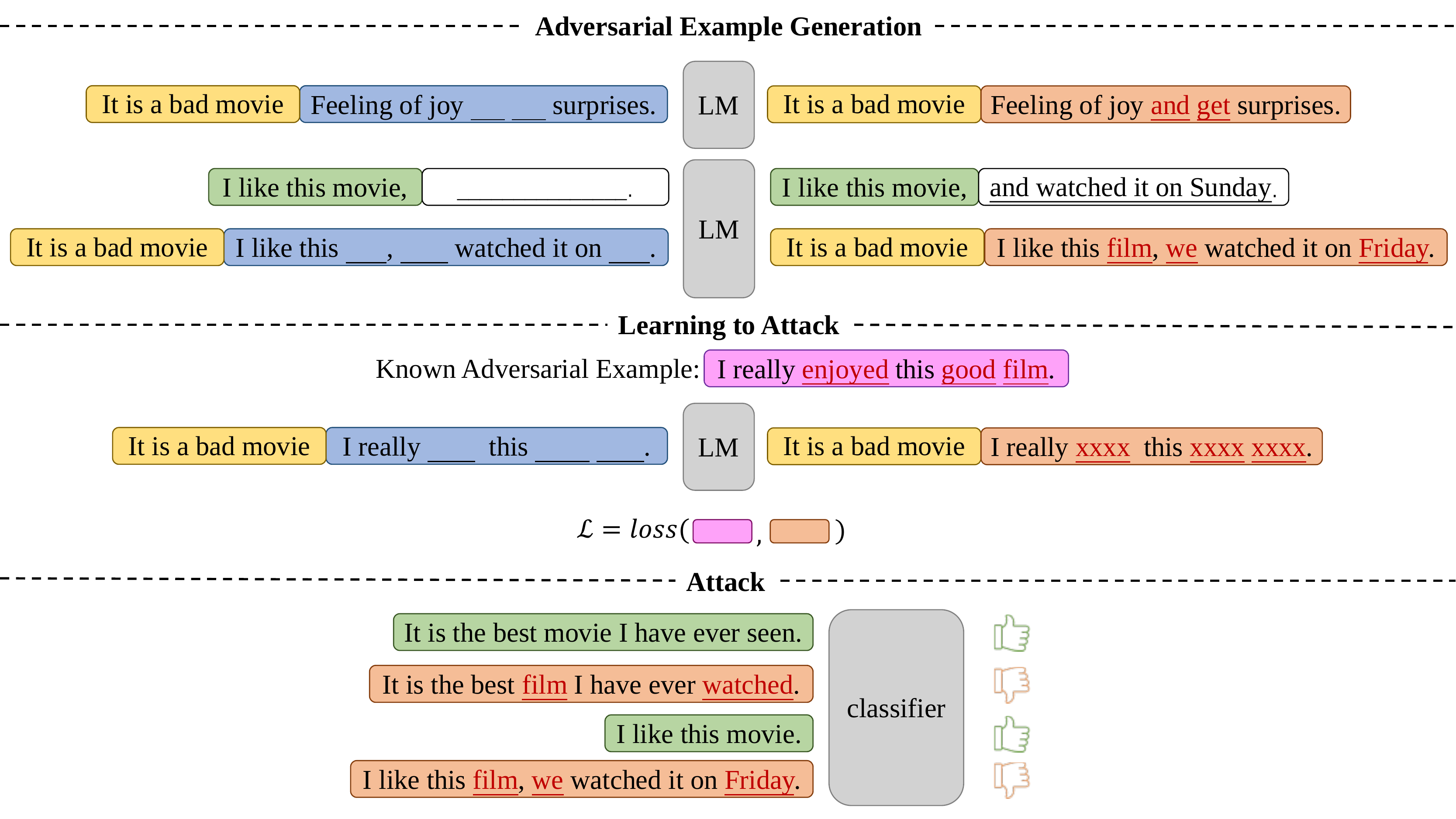}
		\caption{The diagram of our prompt-based attack approach. Block `\img{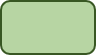}' denotes the original (or normal) example. Block `\img{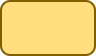}' denotes the trigger text with a malicious purpose.
		Block `\img{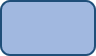}' denotes the masked example with some unfilled slots. Block `\img{Fig/yellow.png}\img{Fig/blue.png}' is a prompt template. Block `\img{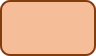}' denotes the adversarial example that we expect the language model to generate which can ``fool'' the downstream classifier. Gesture `\imgg{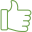}' means the classifier recognizes it with a correct label (positive sentiment) and Gesture `\imgg{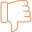}' means a wrong prediction (negative sentiment). The first part presents both word-level and sentence-level attack approches.} 
		\label{fig1}
\end{figure*}

Adversarial example generation is also a sequence generation task that is exactly what a pre-trained language model (PLM) is good at. Thus, we think that PLM can be used to generate high-quality adversarial examples. A new paradigm, prompting, which aims to make full use of PLMs, is driving a second sea change \cite{corr}. Prompting paradigm can reformulate a variety of downstream tasks to the pre-trained tasks with the help of a textual \emph{prompt} which can be seen as keys to ``probe'' the task-related knowledge from PLM. Thus, some questions then arise: if a prompt is maliciously constructed, what will the PLM output? Can prompting paradigm be used in adversarial examples generation and robustness enhancement? To our best knowledge, these questions are still under-explored. 

In this work, we explore the potential of the prompting paradigm in probing the vulnerability of PLMs and enhancing robustness. We know that many downstream models are fine-tuned on PLMs like BERT. Therefore, we utilize the robustness defect of PLMs to generate adversarial examples for such downstream models and design a \textbf{p}romt-based \textbf{at}tack approach (PAT). As shown in the first part of Figure \ref{fig1}, in order to ``probe" the hidden robustness defect of the PLM, PAT first constructs a prompt template which consists of two parts: an input text with some unfilled slots and an additional text with a malicious purpose (we call it a ``trigger"). The trigger text can be inserted in the beginning, middle or end of the masked input. Then, PAT fills in the unfilled slots with a PLM. After removing the trigger text, we get a paraphrased text as the adversarial example candidate. PAT can achieve an impressive attack success rate when feeding the candidates into the victim model. To further improve the attack performance, we use a few known adversarial examples to teach PAT. As the well pre-trained LM is like a talented child, he can be trained to be a skilled hacker via letting him see a few adversarial examples. The diagram of our adversarial example generation approach is shown in the second part of Figure \ref{fig1}. Unlike most search-based attack methods, PAT does not need to iteratively query the victim models. So, it provides a feasible attack solution when the hacker cannot access the detail of the victim model and the number of queries is limited. Experimental results show that PAT can generate more natural, fluent and diverse adversarial examples compared with some search-based attack algorithms.

The purpose of adversarial example generation is to better understand models and aid the construction of robust and reliable deep NLP systems rather than attacking. A natural idea to improve the model is to retrain it with adversarial examples. However, generating adversarial examples can be very time-consuming, especially for a large dataset. We notice that adversarial examples can be implicitly involved in adversarial training and do not need to be really constructed under our prompt-based paradigm. As an adversarial example is generated by decoding the prompt embedding in PAT, we can use the prompt embedding as an alternative to the adversarial example. We fine-tune the embedding space in which similar sample pairs (original and adversarial ones) should stay closer. Experimental results show that our training method can efficiently improve the robustness of the neural network.

In summary, we implemented a preliminary prompting-based method for adversarial example generation and robustness enhancement. It reveals that prompting paradigm has a great potential in better understanding of PLMs and pursuing reliable NLP systems. 

\section{Preliminary}

\paragraph{Prompting}
In recent two years, prompting is popular and leading an effective way to utilize the pre-trained models. In this paradigm, instead of adapting pre-trained LMs to downstream tasks via objective engineering, downstream tasks are reformulated to look more like those solved during the original LM training with the help of a textual prompt~\cite{corr}. For example, when recognizing the emotion of a movie review, ``\textit{It is the best movie I have ever seen}", we may continue with a prompt ``\textit{overall, it is a $\_\_\_$ movie.}", and ask the LM to fill in the blank with an emotion-bearing word. Or if we select a prompt like ``\textit{English: I like this movie. French: $\_\_\_$}", an LM may be able to fill in the blank with a French translation. The advantage of this method is that, given a suite of appropriate prompts, task-related knowledge can be easily probed from LM.
\paragraph{Adversarial Example}
The purpose of an adversarial attack is to perturb a normal input example $x$ to generate an adversarial example $x_{adv}$ for a target model (e.g. a sentiment classifier) so that $x_{adv}$ can mislead the neural network to incorrect predictions. The perturbed example $x_{adv}$ should be semantically preserved for human judge.

\section{Adversarial Example Generation}
As adversarial example generation is a kind of sequence generation task, we believe that the strong generation ability of PLM can be helpful for generating more natural, diverse and fluent adversarial examples. The prompting-based methods have been successfully applied to various NLP tasks \cite{GaoFC20,LiL20} as they can probe the task-related knowledge from PLMs efficiently. So, we wonder if prompting paradigm could be used to probe the robustness defect of PLMs and realize attacks for downstream models. In this paper, we take the first step and propose a prompt-based attack paradigm called PAT. Figure \ref{fig1} presents the diagram of our generation method. In the following section, we will depict how to implement word-level and sentence-level attacks in detail.

\subsection{Word-level Generation}
In this section, we introduce how to realize a prompting-based word-level attack, which aims at crafting adversarial examples via modifying words in the original text. Existing word-level attacks are mainly based on searching substitutions from some pre-given word candidate sets. Our generative attack method prompts a PLM to modify some words in the original input under the effect of a malicious trigger text. Unlike these search-based attacks, our method does not select a substitution from a given set but generates it with the given context. So, it can generate more diverse and fluent sentences. PAT consists of two main steps: prompt construction and candidates generation.

\paragraph{Prompt Construction}
In the classical prompt paradigm, a downstream task will be reformulated into a mask-filling task via applying a \textit{prompt function} $f_{p}$ to the input $x$. The answer of the task lies in the slots filled by the PLM. Taking text classification as an example, given an input $x=$ ``\textit{It is the best movie I have ever seen.}" Then we have a possible prompt: $x_p =f_{p}(x)= ``[x] \textit{ It is a \_\_\_ movie.}"$ where the target prediction of the blank is ``\textit{good}". 

we propose a variant prompt construction method which contains two main steps: masking some positions in $x$ to get $x'$ and concatenating $x'$ with a malicious trigger. To lead the process of mask-filling towards altering the prediction of downstream models, a trigger, which is an additional text containing the attacker's malicious purpose, should be designed. Such triggers can be label-related. Given a normal example $x$, these two steps can be briefly written as:

\begin{enumerate}
    \item  $Generate \ x'= mask(x)$.
    \item $Generate \ a \  prompt$ $x_p=f_{p}(x',label)$.
\end{enumerate}
where $f_p(x',label)$ denotes the operation of concatenating $x'$ with a label-related trigger.


For classification tasks, the trigger can be a simple sentence describing the target label. One of the possible choice of $f_p(x',label)$ for sentiment classification task can be:
\begin{align*}
f_p(x', positive) & = \textit{ It is a bad movie,} [x'] \\
f_p(x', negative) & = \textit{ It is a good movie,} [x']
\end{align*}
For natural language inference task, it aims to learn the relationship between two texts (premise $x_1$ and hypothesis $x_2$): entailment, contradiction or neutral. Attacking is usually conducted on $x_2$. Therefore, we only mask $x_2$ and $f_p(x',label)$ can be designed as:
 \begin{align*}
    f_p(x', entailment) & = [x_1] \textit{ is contradictory with: }[x_2']\\
    f_p(x', contradiction) & = [x_1]\textit{ , implying that: }[x_2']
\end{align*}
The function $f_p(x',neutral)$ can be ``$[x_1]$ \textit{is contradictory with: }$[x_2']$ " or  `` $[x_1]$\textit{ , implying that: }$[x_2']$".


\paragraph{Candidates Generation} 
After prompt construction, we ask a PLM to fill in the blanks in prompt $x_p$.
With the joined effect of the trigger and the original context, PLM may fill in the blank with context-consistent words that can keep the sentence fluent but the semantic embedding will be shifted towards the trigger-guided direction. In order to avoid semantic inversion caused by the triggers, we use the antonym dictionary of WordNet \cite{wordnet} to prevent the generation of antonyms.
After filling in the blanks, we remove the trigger and get the adversarial example candidate. The attacking progress does not need iterative queries of victim models.

\subsection{Sentence-level Generation}
For sentence-level attacks, a natural idea is to insert a succession of blanks to the original input and ask the PLM to fill in these blanks under the guidance of a malicious trigger. However, successive unfilled slots may be filled with semantic-changed phrases owing to the loose constraint of context. Thus, our sentence-level attack is implemented in two steps. We first use the PLM to continue writing a sentence or insert a sentence to the original input $x$. This step can be regarded as a prompting method with a \textit{null} trigger. After getting the new concatenated text $\widetilde{x}$, we reformulate this problem to a word-level attack. So, the outline of sentence-level attack is:
\begin{enumerate}
    \item $\widetilde{x}=insert(x)$.
    \item $Call \  word$-$level\  attack\  with\  \widetilde{x}\  as\  the \ input.$
\end{enumerate}
The purpose of the first step is to add a sentence into the original text $x$ which is semantic-consistent and context-natural. The second step aims to modify text $\widetilde{x}$ so that it can mislead the model's output. To add a natural sentence, GPT2 is a good choice as it's an auto-regressive language model which is trained on large-scale corpora and shows great performance in renewing sentences given a beginning. A simple but effective trick to perform more natural writing is to replace the stop mark at the end of input text with a comma which implies an unfinished sentence.

\begin{figure*}
		\centering
		\includegraphics[scale=0.5]{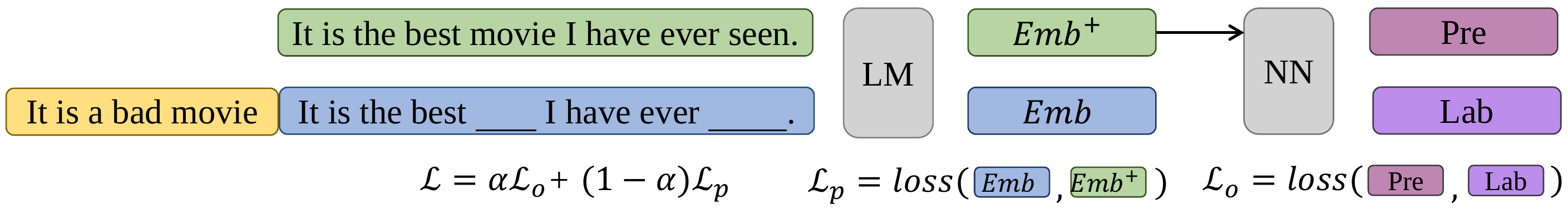}
		\caption{The diagram of our robustness enhancement method. Block `\img{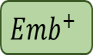}' and `\img{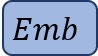}' correspond to the embedding of original text and prompt respectively. Block `\img{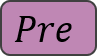}' denotes the prediction of the classifier and `\img{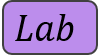}' is the true label.}
		\label{fig2}
\end{figure*}

\subsection{Improve Attack Ability by Learning}\label{sec:attack_learning}
Sometimes using PLMs directly to fill in the slots and generate adversarial examples may not have a satisfactory success rate. We know that the modern PLMs are usually trained on large datasets to learn general-purpose features. Thus, a PLM can be regarded as a gifted child but may not be good at hacking now. We can teach him how to generate adversarial examples by showing him some examples. In the prompting paradigm, a few examples can help PLMs learn the downstream tasks efficiently. So, we first generate a few adversarial examples of the victim model. For each adversarial example, we mask the positions where they have been changed compared to the original sample and add a trigger to it. We ask the PLM to fill in the blank and predict the original token of adversarial example with cross entropy loss. The second part of Figure \ref{fig1} is a diagram of our training method.


\section{Robustness Enhancement}
A natural idea to improve robustness is to add adversarial examples to the training set and retrain the model. However, generating adversarial samples for a large training set can be very time-consuming even if each adversarial example generation takes only a few seconds. 

A PLM usually outputs a sentence embedding which is a key for downstream tasks. Thus, we can fine-tune the embedding space to bring the embedding of original input and its adversarial example closer. As adversarial examples are decoded from prompts in our method, we omit the progress of decoding and directly fine-tune the embeddings of prompts. As shown in Figure \ref{fig2}, for each pair of original input $x$ and its prompt $x_p$, their sentence embeddings are outputted by a LM denoted as $f(x)$. The training objective is to make the two sentence embeddings closer to each other. A feasible loss function can be mean squared error loss:
\begin{equation*}
   \mathcal{L}_{p}= \frac{1}{|\mathcal{X}|}\sum_{x \in \mathcal{X}}(\left\|f(x_p)-f(x)\right\|_2^2)
\end{equation*}

We assume that $\mathcal{L}_{o}$ is the original loss used by the downstream task, then the final training loss can be modified as:
\begin{equation*}
   \mathcal{L}= \alpha \mathcal{L}_{o} + (1-\alpha) \mathcal{L}_{p}
\end{equation*}
where parameter $\alpha$ is used to make a trade-off between the two losses. The first part of the loss aims at maintaining clean accuracy and the second part is for robustness improvement. Our training method can be seen as a kind of adversarial data argumentation, but the adversarial examples are implicitly involved in the process. Since we only need to add a trigger and mask a few positions for each training instance, the cost is very cheap.

\section{Experiments}
\paragraph{Datasets and Target Models}
We conduct experiments on two important NLP tasks (text classification and natural language inference). For text classification, MR \cite{mr} and IMDB \cite{imdb} are sentence-level and document-level classification tasks respectively. SNLI \cite{snli} is the dataset for the natural language inference task: whether the second sentence (hypothesis) can be derived from the first sentence (premise) with entailment, contradiction, or neutral relationship. Target models include two popular deep neural architectures (BiLSTM \cite{bilstm} and BERT \cite{bert}). LSTM is used for studying the transferability of adversarial examples generated by PAT. We used a 1-layer bidirectional LSTM with 150 hidden units, and 300-dimensional pre-trained GloVe word embeddings \cite{glove}. We used the 12-layer based version of BERT model with 768 hidden units, 12 heads, and 110M parameters. 


We use TextFooler \cite{textfooler} as a reference which is a search-based word substitution algorithm. 
To reduce invalid substitutions of TextFooler, we rank the original candidate set generated based on HowNet \cite{hownet} via similarities of word embeddings and keep the top 5 candidates for each replaceable position. Note that if TextFooler uses the original candidate set (50 synonyms) described in the paper, the success rate of attack is larger than 90\%. 

\begin{table*}
	\resizebox{1\textwidth}{!}{
		\centering
		\begin{small}
			\begin{tabular}{lc|cr||crr|crr|crr|crr}
				\toprule
				\multirow{3}*{Dataset} & \multirow{3}*{Model} &\multicolumn{2}{c||}{TextFooler}&\multicolumn{6}{c|}{Word-level}&\multicolumn{6}{c}{Sentence-level}\\
				&&&&\multicolumn{3}{c|}{PAT}&\multicolumn{3}{c|}{PAT*}&\multicolumn{3}{c|}{PAT}&\multicolumn{3}{c}{PAT*}\\
				&&Suc&PPL$\downarrow$ &Suc&PPL$\downarrow$ & Dis&Suc&PPL$\downarrow$ & Dis &Suc&PPL$\downarrow$& Dis &Suc&PPL$\downarrow$ & Dis\\
				\hline
				\multirow{2}*{MR}& BiLSTM & 69.70 & 694.67 & 42.04& 419.26 & 33.33 & 50.96 & 574.82& 30.00 & 46.45 & 90.51& 33.33 & 59.35 & 128.15 & 36.96\\
				&BERT & 48.35 & 600.49 & 43.02 & 395.77& 50.65 & 55.31 & 590.32& 44.44 & 39.37 & 81.57& 46.03 & 57.50 & 108.51& 46.74\\
				\hline
				\multirow{2}*{IMDB}& BiLSTM & 86.24 & 188.58 & 38.46 & 81.22& 12.86 & 55.49 & 147.46& 9.90 & 41.53 & 86.05& 14.47 & 50.27 & 143.54& 13.04\\
				&BERT & 82.63 & 191.90 & 30.27& 93.96& 16.07 & 53.51 & 148.49& 15.15 & 37.97 & 90.35& 21.13 & 52.41 & 140.77& 19.39\\
				\hline
				\multirow{2}*{SNLI}& BiLSTM & 75.16 & 1322.70 & 64.85 & 516.67& 19.44 & 83.64 & 759.99 & 23.91 & 44.07 & 49.02& 34.62 & 55.93 & 50.15&  33.33\\
				&BERT & 69.94 & 1023.13 & 66.29 & 456.09& 27.19 & 84.00 & 602.43& 29.66 & 51.11 & 44.71& 36.23 & 75.56 & 35.90& 37.25\\
				\bottomrule
			\end{tabular}
		\end{small}
 	}
	\caption{Attack results of different methods. Suc is the ratio of successful attacking. PPL is the average language perplexity of generated adversarial examples. A Lower value of PPL indicates higher fluency and naturalness. Dis is the ratio of ours successfully attacking instances which are attacked unsuccessfully by TextFooler. PAT and PAT* use BERT without and with attack learning for attacking respectively.}
	\label{tab:attack_reslults}
\end{table*}

\paragraph{Metrics}
Two metrics are used for evaluating attack methods: \textbf{Suc} and \textbf{PPL}. 
Suc is the attack success rate. PPL of GPT2, Language perplexity, is usually used to evaluate the fluency and naturalness of sentences. For robustness evaluation, we also present clean accuracy (\textbf{Acc}) and robustness accuracy (\textbf{Rob}). Rob is the accuracy of a model under attack which is a more comprehensive indicator compared with Suc. 

\paragraph{Setting} For PAT\footnote{Source code will be released after the anonymous review period.}, we construct 50 prompts with 15\% random masked positions for each instance as a batch.

\subsection{Results of Attacking Methods}
Experimental results of different attack methods on randomly sampled 200 test data are presented in Table \ref{tab:attack_reslults}. PAT* utilizes the BERT which is improved by attack learning in Section \ref{sec:attack_learning} to perform mask-filling while PAT uses the original BERT. We find that:
\begin{itemize}

\item Compared with TextFooler, PAT achieves the lower PPL value on all three data sets and two different models. It indicates that PAT can generate more fluent and natural adversarial examples. 

\item PAT can generate novel and diverse adversarial examples that are different from the adversarial examples generated by synonyms substitution. Table \ref{tab:cases} presents some cases. In the first case, the original input (Org) is ``\textit{The film might have been more satisfying if it had, in fact, been fleshed out a little more instead of going for easy smiles.}" TextFooler performs successful attacking via several synonym substitutions: substituting ``\textit{have}" to ``\textit{experience}", ``\textit{had}" to ``\textit{took}", ``\textit{fact}" to ``\textit{matter}" and ``\textit{easy}" to ``\textit{gentle}". PAT realizes a different paraphrase: replacing ``\textit{The film}" with ``\textit{While it}" and ``\textit{fleshed out}" with ``\textit{thinking going}". The sentiment-related semantic is consistent with a different expression. It also indicates that PAT can perform a phrase-level attack while masked positions are adjacent. In the third case, PAT replaces ``\textit{is}" with a punctuation ``\textit{,}" due to its vocabulary-size generation space.


\item For sentence-level attack, given the original input, PAT can generate a natural sentence which can keep the original task-related semantic and are human-like. For example, given the original input (``\textit{Rates an for effort and a for boring}"), PLM continues to write a natural sentence: ``\textit{it's hard to think of a better way to describe it.}" The close PPL values among word-level and sentence-level examples in Table \ref{tab:attack_reslults} also imply the conclusion.

\item Utilizing a few adversarial examples for attack learning is efficient. Based on the comparisons of PAT and PAT*, we can see that the attack success rate can be improved with more than 10\% using less than 500 training data. The PPL increases at the same time, mainly owing to that PAT* is trained by adversarial examples generated by TextFooler which are less fluent. 

\item Although PAT can not outperform the search-based method in terms of success rate, the diversity of its generation leads to some interesting attacks in the cases that TextFooler fails. As shown in the second case in Table \ref{tab:cases}, PAT attacks successfully via substituting ``\textit{the}" to ``\textit{it}" while TextFooler fails because they are not synonyms. PAT can provide complementary adversarial examples for the traditional attack method. Take MR-BERT as an example, The Dis of 50.65\% means that 50.65\% successfully attacking instances of word-level PAT is attacked unsuccessfully by TextFooler. 

\item PAT performs better in short-text tasks. It can achieve close Suc with TextFooler in MR and SNLI tasks, and even higher Suc in some cases, especially after attack learning. For IMDB, the best performance with the word-level attack using attack training is 53.51\% for BERT while TextFooler achieves 82.63\%. It's mainly due to the position-wise transformer architecture. As we add the trigger at the beginning of the long text, its influence is limited. The impressive results of Suc and PPL on BiLSTM also imply the good transferability of adversarial examples.
\end{itemize}

\paragraph{Influences of Triggers}
We present the word-level attack results using different triggers in Table \ref{tab:triggers_results}. We can see that the attack experiments with different triggers perform differently although the semantics of these triggers are very similar. It indicates the design of prompts is a feature engineering problem worthy of study.

\begin{table}[t]
	\resizebox{0.5\textwidth}{!}{
		\centering
		\begin{small}
			\begin{tabular}{l|c|l|c}
				\toprule
			    Triggers & Suc & Triggers & Suc \\
			    \hline
			    It is a good movie. & \multirow{2}*{43.02} & It is a funny movie. & \multirow{2}*{37.43} \\ 
			    It is a bad movie.&&It is a boring movie.&\\
			    \hline
			    I like the movie so much. & \multirow{2}*{35.71} & I think it is funny. & \multirow{2}*{34.64} \\ 
			    I hate the movie so much. &&I think it is boring.&\\
				\bottomrule
			\end{tabular}
		\end{small}
		}
	\caption{The influences of triggers: word-level attack results using different triggers on MR-BERT of PAT.}
	\label{tab:triggers_results}
\end{table}

\begin{table*}[t]
 	\resizebox{1\textwidth}{!}{
		\centering
		\begin{small}
			\begin{tabular}{l|l}
				\toprule
				\multicolumn{2}{c}{Word-level} \\
				\hline
				Org & The film might have been more satisfying if it had, in fact , been fleshed out a little more instead of going for easy smiles.\\
				Label & 0 (Negative sentiment) $\rightarrow$ 1 (Positive sentiment)\\
				TextFooler & The film might \textbf{experience} been more satisfying if it \textbf{took}, in \textbf{matter}, been fleshed out a little more instead of going for \textbf{gentle} smiles.\\
				PAT & \textbf{while it} might have been more satisfying if it had, in fact, been \textbf{thinking going} a little more instead of going for easy smiles.\\
				\hline
				Org & You watch for that sense of openness, the little surprises.\\
				Label & 1 (Positive sentiment) $\rightarrow$ 0 (Negative sentiment)\\
				TextFooler & \textbf{[Attack failed]}\\
				PAT & You watch for that sense of openness, \textbf{of} little surprises.\\
				\hline
				Org & [x1]: A guy riding a motorcycle near junk cars. [x2]: A man is riding a motorcycle.\\
				Label & 2 (Entailment) $\rightarrow$ 1 (Contradiction)\\
				TextFooler & ... [x2]: A \textbf{man} is riding a \textbf{motorbike}.\\
				PAT & ... [x2]: A \textbf{young ,} riding a motorcycle.\\
				\hline
				\multicolumn{2}{c}{Sentence-level} \\
				\hline
				Org & The film is predictable in the reassuring manner of a beautifully sung holiday carol.\\
				Label & 1 (Positive sentiment) $\rightarrow$ 0 (Negative sentiment)\\
				Continued sentence & but it ' s also one of the funniest movies I' ve ever seen. \\ 
				PAT & The film is predictable in the reassuring manner of a beautifully sung \textbf{Christmas} carol, \textbf{but it' s also one of the best movies I've ever seen.}\\
				\hline
				Org & Rates an for effort and a for boring. \\
				Label & 0 (Negative sentiment) $\rightarrow$ 1 (Positive sentiment)\\
				Continued sentence & it' s hard to think of a better way to describe it.\\
				PAT & Rates an for effort and a for boring, \textbf{it' s hard to think of a better way to like it.}\\
				\bottomrule
			\end{tabular}
		\end{small}
	}
	\caption{Cases of adversarial examples generated by different attacking methods. Org is the original input text. \textbf{Bold font} marks the changed words in adversarial examples.}
	\label{tab:cases}
\end{table*}

\subsection{Results of Defense Methods}
Then, we investigate the effect of our robustness enhancement method. We compare the robustness of models which are original, enhanced with adversarial training (Adv) and our robustness enhancement method. As generating adversarial examples is time-consuming, we only generate adversarial examples for 25\% of the training data and retrain the model via traditional adversarial argumentation. For our method, we construct (input, prompt) pairs for the whole training set. We set $\alpha=0.1$ for MR, SNLI and $\alpha=0.2$ for IMDB.

Experiment results are shown in Table \ref{tab:defense_results}. We find that: our enhancement method always achieves better performances of robustness accuracy for different target models under two attack methods. Besides, our method has a good trade-off between clean accuracy and robustness. It can maintain the clean accuracy with a decrease of less than 2\% and improve robustness accuracy significantly. Compared with conventional adversarial training, our prompt-based training method can improve the robustness accuracy about by 10\% while maintaining high clean accuracy. This shows that our method can fix some inherent vulnerabilities probed by prompt templates.

\begin{table}[t]
		\centering
		\begin{small}
			\begin{tabular}{llr|rr|rr}
				\toprule
				& & & \multicolumn{2}{c|}{TextFooler} & \multicolumn{2}{c}{PAT}\\
				& & Acc & Suc$\downarrow$ & Rob & Suc$\downarrow$ & Rob \\
				\hline
				\multirow{3}*{MR} & BERT & 89.60 & 48.35 & 46.50 & 43.02 & 51.00\\
				& Adv & 88.00 & 40.22 & 52.00 & 43.82 & 49.00\\
				& Ours & 89.50 & 30.56 & 62.50 & 36.87 & 56.50 \\
				\hline
				\multirow{3}*{IMDB} & BERT & 93.68 & 82.63 & 16.50 & 30.27 & 65.00\\
				& Adv & 91.00 & 38.95 & 58.00 & 28.12 & 65.50\\
				& Ours & 92.20 & 26.78 & 67.00 & 12.71 & 80.50\\
				\hline
				\multirow{3}*{SNLI} & BERT & 86.77 & 69.94 & 26.00 & 66.29 & 29.50\\
				& Adv & 82.53 & 52.98 & 39.50 & 73.49 & 22.00\\
				& Ours & 84.81 &41.42 &50.50 &47.33 &44.50\\
				
				\bottomrule
			\end{tabular}
		\end{small}
	\caption{Defense results of different methods. Acc is the clean accuracy on test set. Rob measures the accuracy under attacking.}
	\label{tab:defense_results}
\end{table}

\section{Related Work}
\paragraph{Prompting}
Recently, the ``\textit{pre-train, fine-tune}" procedure is gradually replaced by one called ``\textit{pre-train, prompt, and predict}".  In this way, by selecting the appropriate prompts we can manipulate the model behavior so that the pre-trained LM itself can be used to predict the desired output, sometimes even without any additional task-specific training. The prompt-based approach has shown great potential for many NLP tasks including text classification \cite{GaoFC20}, natural language inference \cite{SchickS21}, summarization \cite{LiL20,DouLHJN21}, dialogue systems \cite{sup-dst-prompt}, etc.

\paragraph{Adversarial Attack}
Various search-based attack algorithms are developed for generating adversarial examples including gradient descent methods \cite{Wang0D021}, genetic algorithm \cite{AlzantotSEHSC18}, particle-swarm-based method \cite{seme}, greedy-based methods \cite{Ren19,textfooler} and BERT-based methods \cite{LiMGXQ20}. Although these methods can achieve a very high attack success rate, they need to iteratively query the victim model. These methods may not be efficient enough to generate adversarial examples for large training data set. Besides, in some cases, the attacker may not be able to access the target model multiple times.

\paragraph{Defense}
Adversarial training is one of the most popular empirical defense methods. \cite{Ren19,LiMGXQ20,Wang0D021} adopt the adversarial examples generated by their attack methods for adversarial training and achieve robustness improvement. Other empirical defense methods like FGWS \cite{MozesSKG21} and DISP \cite{ZhouJCW19} try to detect adversarial examples. FGWS exploits the frequency properties of adversarial word substitutions for the detection of adversarial examples. DISP learns a perturbation discriminator to identify malicious perturbations and block adversarial attacks. However, black-box attack algorithms can still successfully break the defense when the model and detector are considered as a whole.

\section{Conclusion}
In this paper, we design prompting-based approaches for adversarial example generation and robustness enhancement. For attack, although our method (PAT) is not always superior in terms of attack success rate compared to traditional search-based attack algorithms, the strength of our method is that it can generate more natural, fluent and diverse adversarial examples which are not limited to a pre-defined space. Besides, our approach does not require multiple queries to the victim model. Thus, it can be a good choice in some more demanding scenarios where the interaction with the victim model is limited. We also design a prompt-based adversarial training process to improve the model's robustness. This training method does not rely on the process of searching adversarial examples, so it can be applied to large-scale training data. Experiments also show that our robustness enhancement method achieved an impressive improvement in resisting attack.
 
Our work reveals that some robustness defects can be ``probed" by carefully selected prompting templates. It also indicates that the prompting paradigm has a great potential in NLP robustness issues and adds a piece of the puzzle to the map of prompting's capability. We hope the results in this paper can inspire future work 
from multiple perspectives. A promising direction, we think, is that some other safety and security problems (e.g. privacy) of NLP models also can be investigated by the prompting method.

\bibliographystyle{named}
\bibliography{ijcai22}
\end{document}